\documentclass[conference,letterpaper]{IEEEtran}
\IEEEoverridecommandlockouts
\usepackage{cite}
\usepackage{amsmath,amssymb,amsfonts}
\usepackage{algorithmic}
\usepackage{graphicx}
\graphicspath{{images/}}  
\DeclareGraphicsExtensions{.pdf,.png,.jpg} 
\usepackage{textcomp}
\usepackage{float}
\usepackage{placeins}
\usepackage{algorithm}
\usepackage{booktabs}
\usepackage{tgpagella}
\usepackage{xcolor}
\usepackage{colortbl}
\usepackage[colorlinks,urlcolor=blue,linkcolor=blue,citecolor=blue]{hyperref}

\usepackage{geometry}
\begin{document}

\title{\vspace{10pt}Uncertainty-Aware LLM-Guided Policy Shaping for Sparse-Reward Reinforcement Learning
\thanks{Code: \href{https://github.com/USD-AI-ResearchLab/uncertainty-aware-llm-rl}{github.com/USD-AI-ResearchLab/uncertainty-aware-llm-rl}}}

\author{
\IEEEauthorblockN{
Ujjwal Bhatta\IEEEauthorrefmark{1},
Utsabi Dangol\IEEEauthorrefmark{1},
Sumaly Bajracharya\IEEEauthorrefmark{1},
Rodrigue Rizk,
KC Santosh}
\IEEEauthorblockA{
USD AI Research, Department of Computer Science\\
University of South Dakota\\
Vermillion, SD 57069, USA\\
\{ujjwal.bhatta, utsabi.dangol, sumaly.bajracharya\}@coyotes.usd.edu,
\{rodrigue.rizk, kc.santosh\}@usd.edu
}
\thanks{\IEEEauthorrefmark{1}These authors contributed equally to this work.}
}
\maketitle          
\begin{abstract}
Sparse rewards and heterogeneous task sequences remain persistent challenges in Reinforcement Learning (RL), often resulting in slow convergence, weak generalization, and inefficient exploration. We propose Uncertainty-Aware LLM-Guided Policy Shaping (ULPS), a novel framework that integrates a calibrated Large Language Model (LLM) into the RL training loop to provide structured, uncertainty-modulated behavioral guidance. ULPS employs an A*-based oracle to synthesize optimal symbolic trajectories, which are used to fine-tune a BERT-based language model. During training, this model supplies action suggestions whose influence is conditioned on epistemic uncertainty estimated via Monte Carlo (MC) dropout. An entropy-based blending mechanism adaptively balances LLM guidance and the learned policy (via Proximal Policy Optimization, PPO), allowing the agent to prioritize reliable priors while preserving adaptability. We evaluate ULPS on the MiniGrid-UnlockPickup benchmark and observe consistent improvements in success rate, reward efficiency, and sample complexity over unguided, uncalibrated, and standard RL baselines. ULPS achieves more than 9\% improvement in execution accuracy after fine-tuning, requires fewer environment interactions, and yields higher reward AUC. Our results demonstrate that integrating symbolic A* trajectories, pretrained language priors, and uncertainty-aware control offers a principled and effective approach to multi-task reinforcement learning in sparse-reward domains, with potential extensibility to partially observable and multi-agent settings.
% Code available at: \url{https://github.com/conference-llm/conference.git}
\end{abstract}


\section{Introduction}

Despite advances in Reinforcement Learning (RL) for sequential decision-making tasks like games, robotics, and navigation~\cite{Mnih2013PlayingAW,Silver2016MasteringTG,Haarnoja2018SoftAO}, sparse rewards and diverse task sequences remain major challenges, limiting sample efficiency and generalization. In sparse-reward settings, agents receive feedback only after long sequences of correct actions, making exploration inefficient and frequently resulting in reliance on random exploration strategies, requiring thousands of episodes to discover successful trajectories~\cite{Guo2020MemoryBT,Salimans2018LearningMR}. To address these challenges, prior work has explored incorporating external knowledge into the RL process for sparse-reward environments, relying on human expertise to guide agent. For example, agents can be trained to imitate actions that align with human-judged preferences, which accelerates learning in complex tasks~\cite{Christiano2017DeepRL}. This shows considerable success, but they face practical limitations in scaling it to collect human feedback across diverse environments~\cite{Casper2023OpenPA}.

Recent advances in natural language processing show that Large Language Models (LLMs) can reason over multi-step information~\cite{Shalev2024DistributionalRI}, generate actions from textual or visual inputs to guide RL agents, and decompose complex tasks into context-aware subgoals and mid-level plans that can be translated into executable actions~\cite{Du2023GuidingPI,Huang2022LanguageMA,Kwon2023RewardDW}. However, there are difficulties when directly incorporating LLMs into decision-making or learning systems~\cite{Bai2022ConstitutionalAH}. Simply injecting language guidance into RL often leads to over-reliance on uncertain suggestions, degraded stability, or bias toward suboptimal heuristics. LLMs have a high degree of confidence in producing inaccurate or hallucinated outputs~\cite{Sun2025LargeLM}. Language models are prone to overconfidence and may make unreliable suggestions that could impact learning compared to human experts~\cite{Zhou2024RelyingOT}. 
Therefore, estimating the model's uncertainty and figuring out when to trust its judgment are crucial.  A key challenge is to calibrate LLM-based priors and modulate their influence based on uncertainty. 
Shoaeinaeini and Harrison~\cite{Shoaeinaeini2024GuidingRL} introduced a more structured solution to this problem by designing a calibrated RL system guided by LLMs. Their method uses Monte Carlo (MC) dropout~\cite{Gal2015DropoutAA} along with entropy-based policy shaping to adjust how much the agent relies on LLM advice in multi-task settings. 
 
Building on this foundation, we propose a unified framework, \textbf{U}ncertainty-Aware \textbf{L}LM-Guided \textbf{P}olicy \textbf{S}haping (ULPS), that uses an A*-based oracle to fine-tune LLM and then integrates LLM judgements into the Proximal Policy Optimization (PPO)-based RL training loop through an uncertainty-aware mechanism. Inspired by~\cite{Shoaeinaeini2024GuidingRL}, we extend their environment and prompting approach with improved policy combination and uncertainty integration for larger-scale tasks. Our contributions are twofold: a) a framework that combines LLM guidance with RL policies using uncertainty-aware, entropy-weighted blending for adaptive policy shaping, and b) a scalable self-supervised method using A*-generated trajectories to fine-tune BERT for multi-task sequential decision-making. We demonstrate its effectiveness with 99.17\% accuracy, a 9\% improvement over prior models, achieving higher reward efficiency and lower complexity in sparse-reward environments.

\begin{figure*}[tbp]
    \centering
    \vspace{-8pt}
    \includegraphics[width=\linewidth, keepaspectratio]{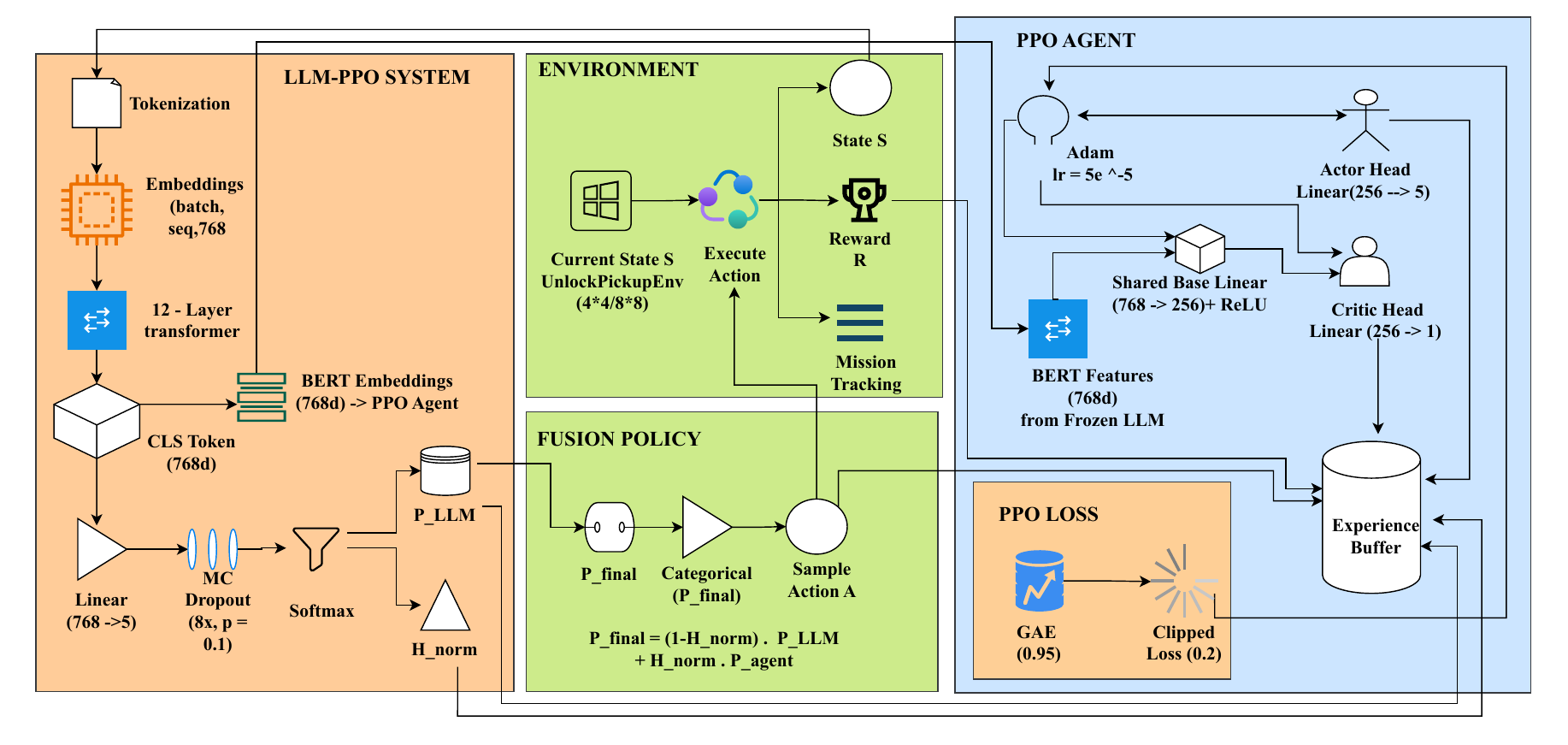}
    \caption{
    Overview of the proposed ULPS framework. An A*-based oracle generates optimal trajectories for BERT fine-tuning. During training, the environment state (S) is encoded and fed into the calibrated LLM. MC dropout estimates uncertainty producing $P_{LLM}$ and $H_{norm}$. The same BERT features are processed by PPO Agent to generate $P_{agent}$. These distributions are fused with entropy-based shaping. The environment executes action (A), returning reward (R) and next state (S'), and the transition tuple (S, A, R, S') is stored for PPO updates. 
    }
    \label{fig:methodology}
\end{figure*}

\section{Methodology}

\subsection{Problem Formulation}
We consider an episodic Markov Decision Process (MDP) $M = \{\mathcal{S}, \mathcal{A}, P, R, \gamma\}$
with sparse rewards, where the agent must solve multiple tasks in MiniGrid-UnlockPickup~\cite{ChevalierBoisvert2023MinigridM}: 
\begin{itemize}
    \item $\mathcal{S}$: State space representing the agent's position, orientation, and environment configuration;
    \item $\mathcal{A}$: Discrete action space, $\mathcal{A}$ = \{turn left, turn right, move forward, pick up, toggle\};
    \item $P$: Transition probability function defining environment; 
    \item \textit{R (S, A)}: Reward function providing sparse feedback for task completion; and 
    \item \textit{$\gamma$}: Discount factor for future rewards. 
\end{itemize} 
Our environment has a sequence of subtasks T = \{$T_1$, $T_2$, $T_3$\}, navigating to and picking up the key, navigating to and unlocking the door, and finally navigating to the goal. The agent receives sparse rewards upon successful completion of each subtask, making traditional RL exploration inefficient, as reward \textit{R} is non-zero for specific transitions (\textit{S}, \textit{A}, \textit{R}, \textit{S'}) where task objective is achieved. \textit{S'} is a new state when agent performs an action \textit{A} while on state \textit{S}. The objective is to learn a policy $P_{\text{final}}(a|s)$ maximizing expected discounted returns. ULPS augments PPO with an LLM-derived prior $P_{\text{LLM}}(a|s)$.

\subsection{System Architecture}
Our model uses a calibrated LLM-based RL system with a PPO agent. The proposed architecture is depicted in Fig. \ref{fig:methodology}.
The agent learns by combining its own policy with the guidance of a language model, modulated by the model's confidence. At the start of each episode, the environment state \( S \) is translated into a textual prompt, which is passed through the fine-tuned BERT model \( T \) times using MC dropout. This produces a distribution over possible actions $P_{\text{LLM}}$, along with an associated entropy \( H \) that captures the model's uncertainty. We then normalize the entropy to obtain \( H_{\text{norm}} \in [0, 1] \), which determines how much weight to assign to the LLM versus the PPO policy. The PPO agent's policy, $P_{\text{agent}}$, is obtained by passing BERT-extracted features through a small actor-critic network.
 The final policy is a convex combination:
\begin{equation}
P_{\text{final}} = (1 - H_{\text{norm}}) \cdot P_{\text{LLM}} + H_{\text{norm}} \cdot P_{\text{agent}}.
\end{equation}

An action \( A \sim P_{\text{final}} \) is then sampled and executed in the environment. The resulting experience tuple (\textit{S}, \textit{A}, \textit{R}, \textit{S'}) is stored in the PPO buffer for future updates. Over time, the PPO agent is updated using these collected trajectories. This adaptive training procedure enables the agent to leverage its own learning and the structured priors embedded in the language model.  The algorithm~\ref{alg:training} summarizes the full process.

\begin{algorithm}[tbp]
\caption{Training with Uncertainty-Aware LLM Guidance}
\label{alg:training} 
\begin{algorithmic}[1]
\REQUIRE Fine-tuned BERT model $\mathcal{B}$, PPO agent, environment $\mathcal{E}$, number of episodes $N$, forward passes $T=8$, dropout rate $p=0.1$
\ENSURE Trained PPO policy $P_{\text{agent}}$, experience buffer for policy updates

\FOR{each episode $e = 1$ to $N$}
    \STATE Initialize $S \leftarrow S_0$
    \WHILE{episode not terminated}
        \STATE $\tau \leftarrow \phi(S)$ \COMMENT{state to text prompt}
        \STATE $P_{\text{LLM}} \leftarrow \frac{1}{T}\sum_{k=1}^{T} \mathcal{B}^{(k)}(\tau;\, p)$ \COMMENT{$T$ stochastic forward passes}
        \STATE $H \leftarrow -\sum_{a \in \mathcal{A}} P_{\text{LLM}}(a) \log P_{\text{LLM}}(a)$
        \STATE 
        $H_{\text{norm}} \leftarrow (H - H_{\min}) / (H_{\max} - H_{\min})$
        \STATE 
        $P_{\text{final}} \leftarrow (1 - H_{\text{norm}}) \cdot P_{\text{LLM}} + H_{\text{norm}} \cdot P_{\text{agent}}$
        \STATE $A \sim P_{\text{final}}$; \quad $R, S' \leftarrow \mathcal{E}(S, A)$
        \STATE buffer $\leftarrow$ buffer $\cup\ \{(S, A, R, S')\}$; \quad $S \leftarrow S'$
    \ENDWHILE
\ENDFOR
\STATE Update PPO using buffer with GAE $\lambda=0.95$, clipping $\epsilon=0.2$
\end{algorithmic}
\end{algorithm}

\subsection{Generating Optimal Trajectories with A*}
We employ an A* planner over the grid-world transition graph to compute optimal action sequences. The A* pathfinding algorithm computes the shortest feasible path while looking for obstacles such as walls and locked doors. We use the Manhattan distance heuristic, defined as $h(\text{pos}, \text{target}) = |\text{pos}_x - \text{target}_x| + |\text{pos}_y - \text{target}_y|$. Each trajectory consists of structured state representations (semantic maps and relative positions) and symbolic actions (e.g., \textit{turn left}, \textit{move forward}, \textit{pickup}).

\subsection{Fine-Tuning a BERT-Based LLM}
We convert state representations into textual prompts (e.g., encoded grid layout) and train a BERT-based next-action predictor using entropy $H_{\text{norm}}$ to obtain a blending weight, and final policy ($P_{\text{final}}$) is determined by combining LLM and PPO policies as described in algorithm~\ref{alg:training}.

\section{Experiments, Results, and Discussion}
\subsection{Implementation details}
The architecture integrates a calibrated LLM, fine-tuned on 21,500 samples using bert-base-uncased embeddings, an input length of maximum of 100 tokens, dropout value of 0.1, and A* pathfinding-generated data, with a PPO agent. PPO agent employs an actor-critic network optimized with AdamW optimizer with a learning rate of 5e-5, batch size of 16, and 5 epochs. The agent is trained for 1,000 episodes, each with a maximum of 50 steps. The policy-shaping mechanism combines probability distributions of both the LLM and the PPO agents based on the normalized entropy of the current state. The entropy coefficient is set to 0.01, value loss coefficient to 0.5, and the GAE-lambda parameter to 0.95. These hyperparameters were selected based on insights from previous research and validated through initial experiments. 
While~\cite{Shoaeinaeini2024GuidingRL} used 4×8 configuration, we adopt 8×4 grid for LLM fine-tuning phase until reaching at least 90\% accuracy and RL training phase with calibrated guidance for 1000 episodes in the 4×4 and 8×8 environments.

After every 50 episodes, PPO updates are performed using clipped policy gradients, value loss, and entropy regularization. This exhibits how uncertainty-calibrated LLMs can guide RL agents, with a smooth transition from LLM guidance to learned policy control as training progresses.

\begin{figure}[tbp]
  \centering
  \includegraphics[width=\linewidth,height=5.8cm,keepaspectratio]{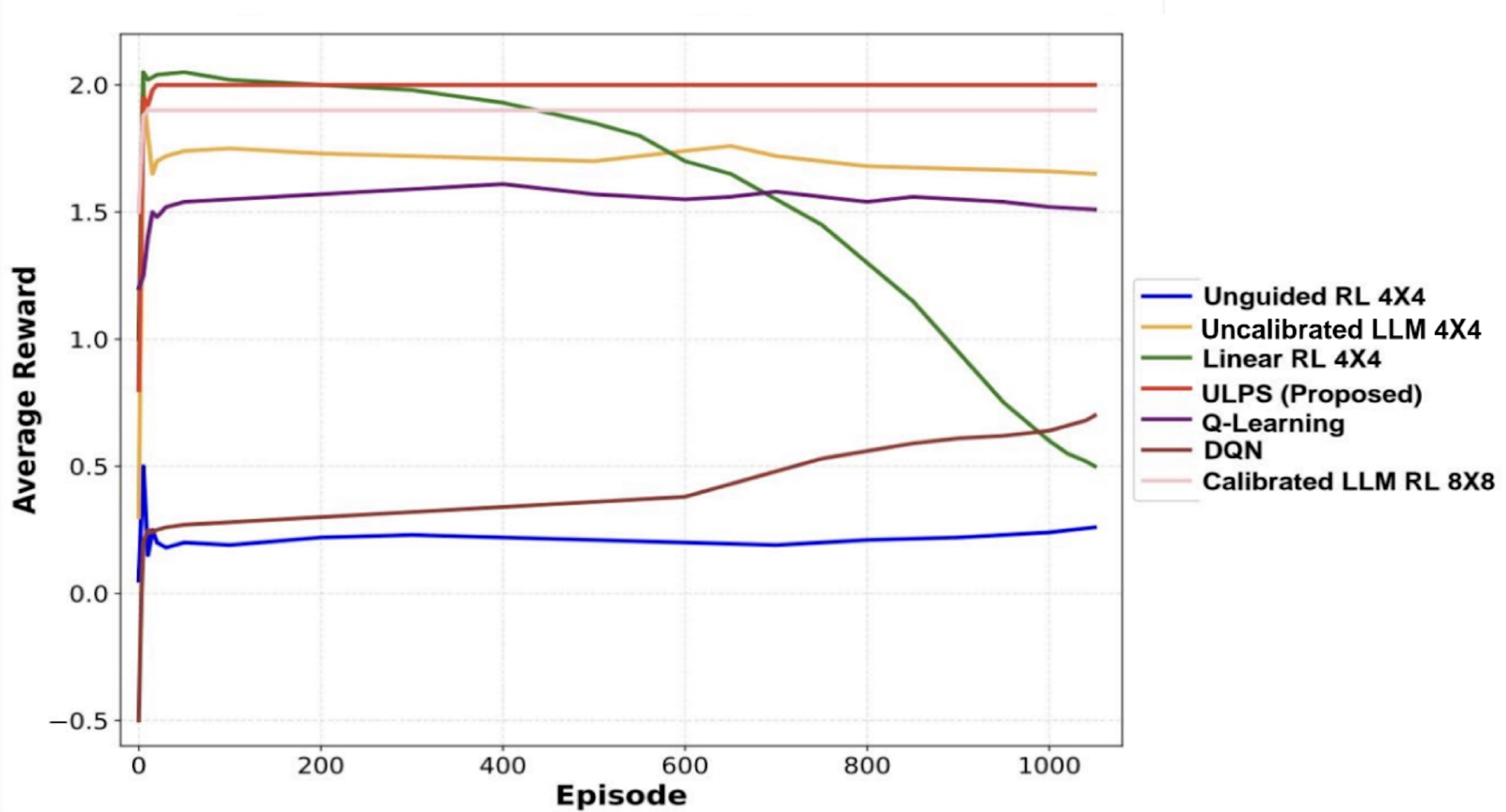}
  \caption{Training performance comparison showing average reward. Our model shows a significantly higher and stable reward trajectory than other baselines. Traditional RL methods like Q-Learning, DQN, and unguided RL show slower learning and lower final rewards. Uncalibrated LLM improves performance but remains less effective than calibrated version.}
  \label{fig:avgReward}
\end{figure}

\begin{figure}[tbp]
  \centering
  \includegraphics[width=\linewidth,height=5.8cm,keepaspectratio]{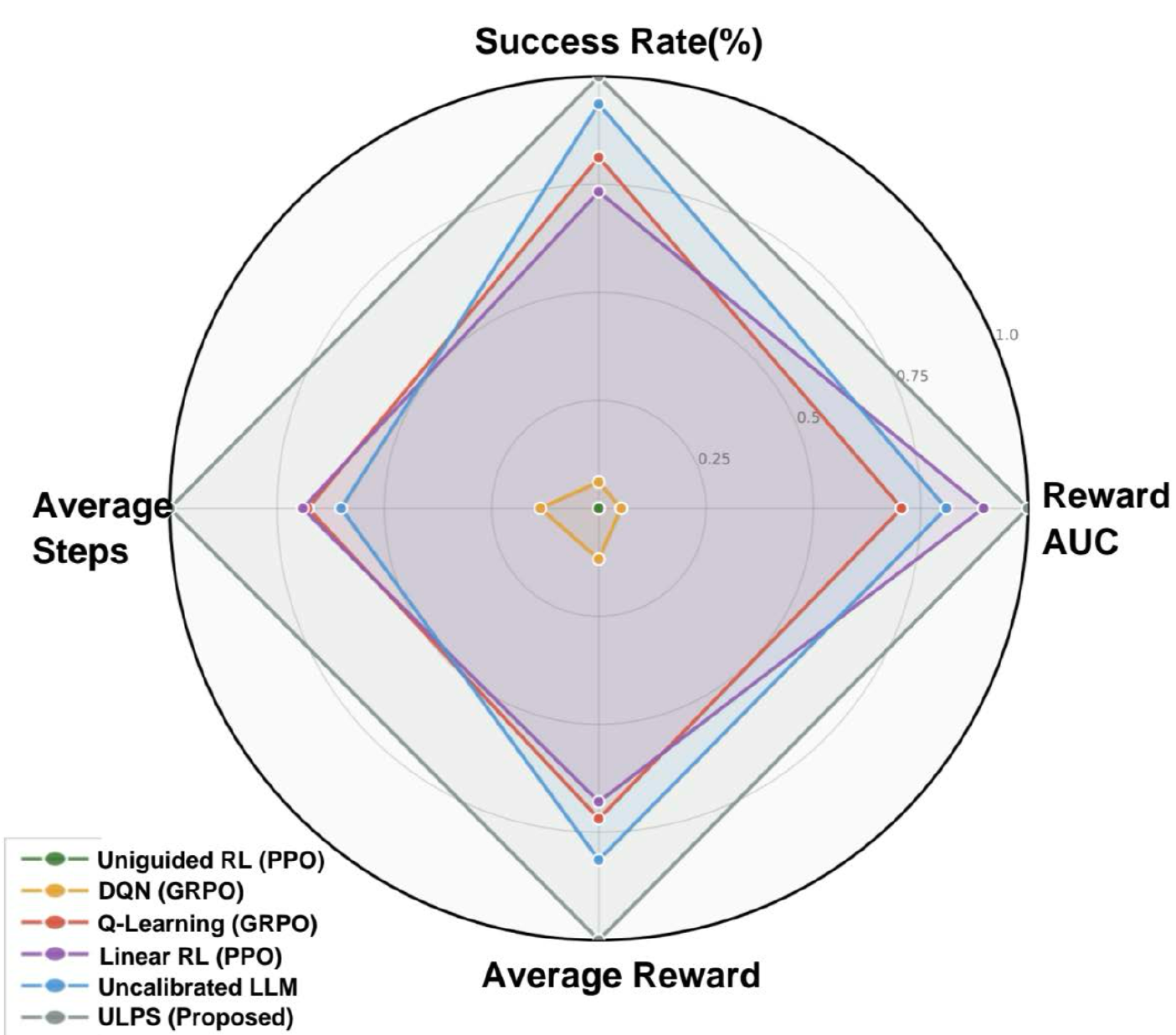}
  \caption{Model comparison based on reward, success rate, steps, and AUC. Our calibrated model achieves the highest scores across all metrics, indicating superior performance. The uncalibrated LLM performs better than traditional methods but falls short of the calibrated model due to its lack of uncertainty awareness.}
  \label{fig:radarChart}
\end{figure}

\subsection{Environment and Reward Structure}
We evaluate our proposed ULPS framework on MiniGrid-UnlockPickup~\cite{ChevalierBoisvert2023MinigridM}, a sparse-reward benchmark for sequential multitask RL. The LLM oracle is trained in an 8×4 environment, while RL is evaluated on 4×4 and 8×8 settings~\cite{FaramaMinigridUnlockPickup2025}. The UnlockPickup environment consists of picking up a key, unlocking a door, and reaching the goal. The observation space includes a 7×7 view, and the action space $A_{t}$ is discrete with five actions: 0 (turn left), 1 (turn right), 2 (move forward), 3 (pick up), and 5 (toggle). The key, door, and goal are located at $(w-2,1)$, $(w-2,h-2)$, and $(w-1,h-2)$ respectively, where $w$ and $h$ denote width and height of the environment. Rewards are based on task performance. The reward function assigns 0.5 for key pickup, 0.5 for door opening, and 0.2 for reaching the goal, with a penalty of $-0.02$ for invalid actions. An additional bonus is added when the goal is reached, where $\textit{AdditionalBonus} = 1 - (\textit{stepscount} / \textit{maxsteps})$. The formula for reward is given by $\textit{Reward} = \textit{KeyPickup} + \textit{OpenDoor} + \textit{ReachGoal} + \textit{AdditionalBonus} + \textit{CumulativePenalty}$.

 The environment ends either upon reaching the goal or exceeding maximum steps (50). This setup shows there is a balanced challenge for exploration and execution of the sequential tasks. The textual prompt format provides a description of the environment at each step with agent locations, orientations, and goals. All experiments were carried out on a fixed random seed (42).

\begin{table}[tbp]
\centering
\caption{Comparison of model calibration and performance metrics. The proposed model achieves higher fine-tuning accuracy, lower bs, and perfect da, indicating superior prediction confidence and reliability compared to prior work.}
\renewcommand{\arraystretch}{1.2}
\setlength{\tabcolsep}{3pt}
\resizebox{0.8\columnwidth}{!}{%
\begin{tabular}{lcccc}
\toprule
\textbf{Model} & \textbf{Acc. (\%)} & \textbf{ECE} & \textbf{BS} & \textbf{DA} \\
\midrule
Shoaeinaeini and Harrison, 4×4~\cite{Shoaeinaeini2024GuidingRL} & 90.00 & 0.15 & 0.20 & 0.80 \\
Shoaeinaeini and Harrison, 3×3~\cite{Shoaeinaeini2024GuidingRL} & 93.00 & \textbf{0.14} & 0.19 & 0.75 \\
Unguided RL (PPO), 4×4 & 99.17 & 0.35 & 0.33 & 0.20 \\
Uncalibrated LLM, 4×4 & 99.17 & 0.33 & 0.31 & 0.33 \\
\midrule
\textbf{Proposed model (Ours)} & \textbf{99.17} & 0.20 & \textbf{0.06} & \textbf{1.00} \\
\bottomrule
\end{tabular}
}
\label{tab1}
\end{table}

\subsection{Results}
Our model's performance for a 4×4 Mini-grid environment has achieved an accuracy of 99.17\% after fine-tuning, outperforming~\cite{Shoaeinaeini2024GuidingRL}, which has 90\% accuracy for 21,500 states as shown in Table \ref{tab1}. This accuracy measures how often our fine-tuned LLM picks the same action as the oracle for a given state.

We employed three standard metrics. Brier Score (BS) measures probabilistic prediction accuracy: $BS = \frac{1}{N} \sum_{i=1}^{N} (p_i - y_i)^2$, where $p_i$ is the predicted probability and $y_i$ is the actual outcome. Expected Calibration Error (ECE) quantifies calibration: $ECE = \sum_{m=1}^{M} \frac{|B_m|}{N} |acc(B_m) - conf(B_m)|$, where $B_m$ represents samples in bin $m$. Discrimination analysis (DA) is measured via AUC-ROC: $AUC = \int_{0}^{1} \mbox{TPR}(\mbox{FPR}^{-1}(t))\, dt$ where \mbox{TPR} and \mbox{FPR} are true and false positive rates, respectively. The evaluation framework uses ECE with a 10-bin calculation, and DA  with a 0.5 threshold.

Our model achieved superior performance with BS of 0.06 (vs. 0.20), DA of 1.0 (vs. 0.8), and ECE of 0.20, indicating better-aligned probabilistic predictions, reduced overconfidence, while higher DA shows model's ability to separate correct and incorrect action predictions.

The experiments performed on different minigrid environment sizes using calibrated LLM, uncalibrated LLM, unguided RL, linear RL, Q-Learning, and DQN have shown several important insights, as shown in Table \ref{tab2} and Fig. \ref{fig:avgReward}. Our proposed model (calibrated LLM in a 4×4 environment) significantly outperforms all other approaches with the highest reward Area Under the Curve (AUC) of 2055.08 and the least average steps to the goal (7.24). The average reward is higher than in other experiments, while the total steps taken are only 7286 for 1000 episodes. Comparatively, the 4×4 uncalibrated LLM showed slightly reduced performance (94.00\% success, 1706.43 AUC, 18.39 steps), while traditional RL methods exhibited greater limitations: Q-Learning attained 82.40\% success and 16.19 steps (1515.71 AUC), and DQN managed 11.60\% success with 31.66 steps (317.46 AUC). The 4×4 Unguided RL baseline performed poorest (5.90\% success, 35.54 steps, 221.31 AUC), emphasizing the value of guided exploration.

\begin{table*}[tbp]
\centering
\caption{Ablation study and performance comparison of various rl methods in 4×4 and 8×8 MiniGrid UnlockPickup environments. 
The proposed ULPS model outperforms traditional rl baselines, including q-learning, dqn, and uncalibrated llm variants, demonstrating superior sample efficiency, reward accumulation, and task success.}

\begin{tabular}{llcccccc}
\toprule
\textbf{Environment} & \textbf{Methods} & \textbf{Reward} & \textbf{Success} & \textbf{Avg. Steps} & \textbf{Avg} & \textbf{Total} & \textbf{Total} \\
&  & \textbf{AUC} & \textbf{Rate (\%)} & \textbf{to Goal} & \textbf{Reward} & \textbf{Wins} & \textbf{Steps} \\

\midrule
4×4 & Linear RL (PPO)                  & 1865.57 & 74.90 & 15.84 & 1.45 & 749 & 24412 \\
&  Uncalibrated LLM           & 1706.43 & 94.00 & 18.39 & 1.69 & 940 & 20284 \\
&  Unguided RL (PPO)                & 221.31  & 5.90  & 35.54 & 0.24 & 59  & 49147 \\
& Q-Learning (GRPO)                & 1515.71 & 82.40 & 16.19 & 1.52 & 824 & 22142 \\
&  DQN (GRPO)                        & 317.46  & 11.60 & 31.66 & 0.45 & 116 & 47873 \\
\midrule
& \textbf{Proposed model (Ours)} & \textbf{2055.08} & \textbf{99.90} & \textbf{7.24} & \textbf{2.02} & \textbf{999} & \textbf{7286} \\
\midrule
8×8 &  Linear RL (PPO)                  & -290.0 & 0.1 & 49.99 & -0.29 & 1 & 49993 \\
&  Uncalibrated LLM           & 1113.77 & 72.3 & 38.57 & 1.12 & 723 & 38568 \\
&  Unguided RL (PPO)                & -358.18  & 0.0  & 50.0 & -0.36 & 0  & 50000 \\
&  Q-Learning (GRPO)                 & 627.05 & 41.4 & 40.78 & 0.63 & 414 & 40780 \\
&  DQN (GRPO)                        & 135.5  & 3.0 & 47.89 & 0.14 & 3 & 47890 \\
\midrule
& \textbf{Proposed model (Ours)} & \textbf{1886.80} & \textbf{99.70} & \textbf{15.37} & \textbf{1.89} & \textbf{997} & \textbf{15478} \\
\bottomrule
\end{tabular}
\label{tab2}
\end{table*} 

The reward AUC is computed as $\textit{AUC}_{\textit{reward}} \approx \sum_{i=1}^{n-1} \left( (\textit{avg\_reward}_i + \textit{avg\_reward}_{i+1})/{2}\right) \cdot (\textit{ep}_{i+1} - \textit{ep}_i)$. Success rate is defined as ${\sum_{i=1}^{n} \textit{goal}_i}/{n} \times 100$, average steps to goal as ${\sum_{i=1}^{m} \textit{length}_i}/{m}$, and average reward as ${\sum_{i=1}^{n} \textit{reward}_i}/{n}$. Total wins and total steps are simply $\sum_{i=1}^{n} \textit{goal}_i$ and $\sum_{i=1}^{n} \textit{length}_i$, respectively. The $\textit{goal}_i \in \{0, 1\}$ indicates whether the goal was reached in episode $i$, $n$ is the total number of episodes, $m$ is the number of successful episodes ($\textit{goal}_i = 1$), $\textit{length}_i$ is the number of steps taken in episode $i$, $\textit{reward}_i$ is the reward received in episode $i$, and $\textit{ep}_i$ is the episode index.

As depicted in Table~\ref{tab3}, all models reached a near perfect success rate and usually converged in about 7 steps, except for a dropout rate of 0.05 and 4 forward passes. The highest reward (2055.88) came from a dropout rate of 0.2 with 12 forward passes. However, using a dropout rate of 0.1, 8 forward passes gave nearly the same reward (2055.08) while being computationally cheaper. This means that although larger settings can push performance slightly higher, more moderate settings often provide a better balance between accuracy and efficiency.

\begin{table}[tbp]
\centering
\caption{The effects of varying dropout rates and forward passes. The metrics evaluated include average steps to goal, and reward auc, providing how these hyperparameters influence model's ability.}

\renewcommand{\arraystretch}{0.92}
\setlength{\tabcolsep}{3pt}

\begin{tabular}{cccc}
\toprule
\textbf{Dropout} & \textbf{Passes} & \textbf{Avg. Steps} & \textbf{AUC} \\
\midrule
0.05 & 4  & 50   & 599.46  \\
0.05 & 8  & 7.0  & 2054.36 \\
0.05 & 12 & 7.0  & 2054.72 \\
0.1  & 4  & 7.4  & 2047.54 \\
\textbf{0.1} & \textbf{8} & \textbf{7.24} & \textbf{2055.08} \\
0.1  & 12 & 7.4  & 2055.72 \\
0.2  & 4  & 7.0  & 2054.30 \\
0.2  & 8  & 7.0  & 2055.10 \\
0.2  & 12 & 7.0  & 2055.88 \\
\bottomrule
\end{tabular}
\label{tab3}
\end{table}

\subsection{Ablation Study}
When comparing PPO, uncalibrated LLMs, and our combined model in Table~\ref{tab2}, clear differences emerge. In 4×4 environment, PPO alone achieved 74.9\% success, but it needed many steps, while uncalibrated LLM reached 94\%, but was still inefficient. Our RL + LLM model achieved 99.9\% success with fewer steps. In an 8×8 environment, PPO alone failed almost completely, and the uncalibrated LLM reached 72.3\%, while our model achieved 99.7\% success with much lower cost. This shows that combining both components with uncertainty awareness works better than using either one alone.
\subsection{Discussion}
 Our experimental results demonstrate fundamental advances in LLM-guided RL. The uncertainty-aware calibration mechanism is critical for achieving high reliability and exploration efficiency. While both calibrated models (4×4, 8×8) achieved remarkable success rates exceeding 99\%, the 4×4 model's lower step count (7.24 vs 15.37) reveals that proper confidence calibration enables more optimal path planning and successfully mitigates over-exploration problem common in traditional approaches. Comparative analysis exposes clear limitations in existing methods. Unguided RL and DQN exhibited 4 to 5 times higher steps despite lower success rates. This performance hierarchy, visually confirmed in a radar chart (Fig.~\ref{fig:radarChart}), strongly supports our hypothesis that LLM guidance provides crucial priors for efficient exploration.

\subsection{Comparative Analysis of Calibrated LLM Performance}
\subsubsection{MC Dropout Computational Cost}
Our uncertainty estimation requires 8 forward passes through the BERT model per action. This introduces an 8$\times$ computational overhead compared to single-pass inference, showing notable per-action cost increase.

\subsubsection{Sample Efficiency vs Computational Trade-off}
The calibrated LLM with MC dropout increases per-action computation but requires only 7,286 total steps compared to 49,147 steps for unguided RL (86\% reduction in environment interactions). This suggests that even if there is a higher per-step computational cost, overall training efficiency is improved due to the reduced exploration.

\subsubsection{Environment Complexity Scaling}
While in the 8×8 environment, it requires 15.37 steps with 99.70\% success, and in the 4×4 environment, it achieves 99.90\% success with 7.24 steps, the increase in average steps for the larger environment shows that computational overhead scales approximately linearly with environment complexity.

\subsubsection{Comparison with 4×4 Linear RL using PPO}
Linear RL shows reward AUC 1865.57 and requires nearly twice as many steps (15.84 vs 7.24). The main problem is it starts with full LLM control and gradually hands control to the agent as training progresses, while our context-aware calibration provides the right support when it’s needed.

\subsubsection{Comparison with 4×4 Uncalibrated LLM}
Since MC dropout is not used, the uncalibrated model often shows overconfidence in ambiguous situations and under-confident guidance in straightforward situations, leading to nearly double the average steps to the goal (18.39 vs 7.24). This difference supports our main idea that accurately estimating confidence is also as important as making correct predictions for successful LLM-guided RL.

\subsubsection{Comparison with 4×4 Unguided RL using PPO}
Unguided RL performs poorly with a minimal reward AUC of 221.31 and a success rate of 5.90\%. This is due to the difficulty of learning in sparse-reward environments with strictly ordered subtasks, where the uninformed exploration leads to inefficient and penalized action sequences. 

\subsubsection{Comparison with 4×4 Q-Learning using Group Relative Policy Optimization (GRPO)}
Q-learning with GRPO~\cite{Shao2024DeepSeekMathPT} shows a reward AUC of 1515.71 and a success rate of 82.40\%. It lacks consistency due to its reliance on a Q-table and the Markov property, struggling with temporal dependencies and linguistic understanding in sequential tasks which results in poor policy development, requiring nearly double the steps (16.19 vs. 7.24), highlighting the benefit of combining semantic reasoning with RL.

\subsubsection{Comparison with DQN using GRPO}
DQN with GRPO has reward AUC of 317.46 and a success rate of 11.60\%, showing it struggles with sequential tasks. Although we implemented improvements like Double DQN, dueling networks, and prioritized replay, it needs more training data for ordered task. Without guidance, it wastes time exploring unhelpful areas, explaining why it takes 47,873 steps but succeeds only 116 times.

\section{Conclusion}
In this work, we introduced ULPS, a unified framework that integrates calibrated language-model priors, symbolic A* guidance, and uncertainty-aware policy for sparse rewards RL, which provides reliable, interpretable action suggestions that enhance exploration while preserving the adaptability of PPO. Our entropy-based blending mechanism ensures stable training and mitigates over-reliance on uncertain LLM outputs. Empirical results on the MiniGrid UnlockPickup benchmark demonstrate that ULPS improves success rate, sample efficiency, and reward AUC compared with unguided, uncalibrated, Q-learning, and DQN. The calibrated LLM component improves symbolic action accuracy by more than 9\% after fine-tuning and consistently accelerates convergence across training regimes. These findings highlight the effectiveness of combining symbolic planning, pretrained language priors, and uncertainty estimation in a principled RL pipeline. Looking forward, our framework exhibits strong potential for scaling to partially observable scenarios and multi-agent coordination environments. Further exploration of hierarchical prompting, multimodal representations, and tighter integration between planning and language-model reasoning represents a promising direction for advancing robust, generalizable RL systems.

\section*{Acknowledgment}
This work was supported by the National Science Foundation under Grant No. \href{https://www.nsf.gov/awardsearch/showAward?AWD_ID=2346643}{\#2346643}, the U.S. Department of Defense under Award No. \href{https://dtic.dimensions.ai/details/grant/grant.14525543}{\#FA9550-23-1-0495}, and the U.S. Department of Education under Grant No. P116Z240151.
Any opinions, findings, conclusions or recommendations expressed in this material are those of the author(s) and do not necessarily reflect the views of the National Science Foundation, the U.S. Department of Defense, or the U.S. Department of Education.

\bibliographystyle{IEEEtran}
\bibliography{ref}

@article{Mnih2013PlayingAW,
  title={Playing Atari with Deep Reinforcement Learning},
  author={Volodymyr Mnih and Koray Kavukcuoglu and David Silver and Alex Graves and Ioannis Antonoglou and Daan Wierstra and others},
  journal={ArXiv},
  year={2013},
  volume={abs/1312.5602}
}

@article{Silver2016MasteringTG,
  title={Mastering the game of Go with deep neural networks and tree search},
  author={David Silver and Aja Huang and Chris J. Maddison and Arthur Guez and Laurent Sifre and George van den Driessche and others},
  journal={Nature},
  year={2016},
  volume={529},
  pages={484-489},
}

@article{Haarnoja2018SoftAO,
  title={Soft Actor-Critic: Off-Policy Maximum Entropy Deep Reinforcement Learning with a Stochastic Actor},
  author={Tuomas Haarnoja and Aurick Zhou and P. Abbeel and Sergey Levine},
  journal={ArXiv},
  year={2018},
  volume={abs/1801.01290},
}

@inproceedings{Guo2020MemoryBT,
  title={Memory Based Trajectory-conditioned Policies for Learning from Sparse Rewards},
  author={Yijie Guo and Jongwook Choi and Marcin Moczulski and Shengyu Feng and Samy Bengio and Mohammad Norouzi and others},
  booktitle={Neural Information Processing Systems},
  year={2020},
}

@article{Salimans2018LearningMR,
  title={Learning Montezuma's Revenge from a Single Demonstration},
  author={Tim Salimans and Richard J. Chen},
  journal={ArXiv},
  year={2018},
  volume={abs/1812.03381},
}

@article{Christiano2017DeepRL,
  title={Deep Reinforcement Learning from Human Preferences},
  author={Paul Francis Christiano and Jan Leike and Tom B. Brown and Miljan Martic and Shane Legg and Dario Amodei},
  journal={ArXiv},
  year={2017},
  volume={abs/1706.03741},
}

@article{Casper2023OpenPA,
  title   = {Open Problems and Fundamental Limitations of Reinforcement Learning from Human Feedback},
  author  = {Stephen Casper and Xander Davies and Claudia Shia and Thomas Krendl Gibert and Jeremy Scherrer and Javier Rando and others},
  journal = {arXiv preprint arXiv:2307.15217},
  year    = {2023}
}

@article{Shalev2024DistributionalRI,
  title={Distributional reasoning in LLMs: Parallel reasoning processes in multi-hop reasoning},
  author={Yuval Shalev and Amir Feder and Ariel Goldstein},
  journal={ArXiv},
  year={2024},
  volume={abs/2406.13858}
}

@inproceedings{Du2023GuidingPI,
  title={Guiding Pretraining in Reinforcement Learning with Large Language Models},
  author={Yuqing Du and Olivia Watkins and Zihan Wang and C{\'e}dric Colas and Trevor Darrell and P. Abbeel and others},
  booktitle={International Conference on Machine Learning},
  year={2023}
}

@article{Huang2022LanguageMA,
  title={Language Models as Zero-Shot Planners: Extracting Actionable Knowledge for Embodied Agents},
  author={Wenlong Huang and P. Abbeel and Deepak Pathak and Igor Mordatch},
  journal={ArXiv},
  year={2022},
  volume={abs/2201.07207}
}

@article{Bai2022ConstitutionalAH,
  title   = {Constitutional AI: Harmlessness from AI Feedback},
  author  = {Yuntao Bai and Saurav Kadavath and Sandipan Kundu and Amanda Askell and Jackson Kernion and Andy Jones and others},
  journal = {arXiv preprint arXiv:2212.08073},
  year    = {2022}
}

@article{Sun2025LargeLM,
  title={Large Language Models are overconfident and amplify human bias},
  author={Fengfei Sun and Ningke Li and Kailong Wang and Lorenz Goette},
  journal={ArXiv},
  year={2025},
  volume={abs/2505.02151},
}

@article{Zhou2024RelyingOT,
  title={Relying on the Unreliable: The Impact of Language Models' Reluctance to Express Uncertainty},
  author={Kaitlyn Zhou and Jena D. Hwang and Xiang Ren and Maarten Sap},
  journal={ArXiv},
  year={2024},
  volume={abs/2401.06730},
}

@article{Shoaeinaeini2024GuidingRL,
  title={Guiding Reinforcement Learning Using Uncertainty-Aware Large Language Models},
  author={Maryam Shoaeinaeini and Brent Harrison},
  journal={2025 IEEE 7th International Conference on Trust, Privacy and Security in Intelligent Systems, and Applications (TPS-ISA)},
  year={2024},
  pages={363-371},
}

@inproceedings{Gal2015DropoutAA,
  title={Dropout as a Bayesian Approximation: Representing Model Uncertainty in Deep Learning},
  author={Yarin Gal and Zoubin Ghahramani},
  booktitle={International Conference on Machine Learning},
  year={2015},
}

@article{ChevalierBoisvert2023MinigridM,
  title={Minigrid \& Miniworld: Modular \& Customizable Reinforcement Learning Environments for Goal-Oriented Tasks},
  author={Maxime Chevalier-Boisvert and Bolun Dai and Mark Towers and Rodrigo de Lazcano and Lucas Willems and Safem Lahlou and others},
  journal={ArXiv},
  year={2023},
  volume={abs/2306.13831},
}

@article{Shao2024DeepSeekMathPT,
  title={DeepSeekMath: Pushing the Limits of Mathematical Reasoning in Open Language Models},
  author={Zhihong Shao and Peiyi Wang and Qihao Zhu and Runxin Xu and Junxiao Song and Xiao Bi and others},
  journal={ArXiv},
  year={2024},
  volume={abs/2402.03300},
}

@article{Kwon2023RewardDW,
  title={Reward Design with Language Models},
  author={Minae Kwon and Sang Michael Xie and Kalesha Bullard and Dorsa Sadigh},
  journal={ArXiv},
  year={2023},
  volume={abs/2303.00001},
}

@misc{FaramaMinigridUnlockPickup2025,
  author       = {{Farama Foundation}},
  title        = {{MiniGrid-UnlockPickup-v0 Environment}},
  year         = {2025},
  note         = {MiniGrid Documentation}
}
\end{document}